%% file: main.tex
\documentclass[11pt,a4paper]{article}
\usepackage{times,latexsym}
\usepackage{url}
\usepackage{multirow}
\usepackage{amssymb}
\usepackage{graphicx}
\usepackage{algorithm,algcompatible,amsmath}
\usepackage{mathtools}
\usepackage{xcolor}
\usepackage{enumitem}
\usepackage{marvosym}
\usepackage{textcomp}
\usepackage{array}

\usepackage[T1]{fontenc}

\usepackage[acceptedWithA]{tacl2018v2}

\usepackage{xspace,mfirstuc,tabulary}

\newif\iftaclinstructions
\taclinstructionsfalse 
\iftaclinstructions
\renewcommand{\confidential}{}
\renewcommand{\anonsubtext}{(No author info supplied here, for consistency with
	TACL-submission anonymization requirements)}
\newcommand{\instr}
\fi

\iftaclpubformat 

\else

\fi

\title{Densely Connected Graph Convolutional Networks for Graph-to-Sequence Learning}

\author{ Zhijiang Guo$^{1}$\thanks{$^{*}$ Equally Contributed.} ~, Yan Zhang$^{1}$$^{*}$, Zhiyang Teng$^{1,2}$, Wei Lu$^{1}$ \\
	$^{1}$Singapore University of Technology and Design \\
	8 Somapah Road, Singapore, 487372 \\
	$^{2}$School of Engineering, Westlake University, China \\
	\texttt{\{zhijiang\_guo,yan\_zhang,zhiyang\_teng\}@mymail.sutd.edu.sg}\\
	\texttt{tengzhiyang@westlake.edu.cn,  luwei@sutd.edu.sg}
}

\date{}

\begin{document}
	\maketitle

    

\begin{abstract}
We focus on graph-to-sequence learning, which can be framed as transducing graph structures to sequences for text generation. To capture  structural information associated with graphs, we  investigate the problem of encoding graphs using graph convolutional networks (GCNs).Unlike various existing approaches where shallow architectures were used for capturing local structural information only, we introduce a dense connection strategy, proposing a novel Densely Connected Graph Convolutional Networks (DCGCNs). Such a deep architecture is able to integrate both local and non-local features to learn a better structural representation of a graph. Our model outperforms the state-of-the-art neural models significantly on AMR-to-text generation and syntax-based neural machine translation.

\end{abstract}

\section{Introduction}
\label{sec:1}

\begin{figure}
    \centering
    \includegraphics[scale=0.26]{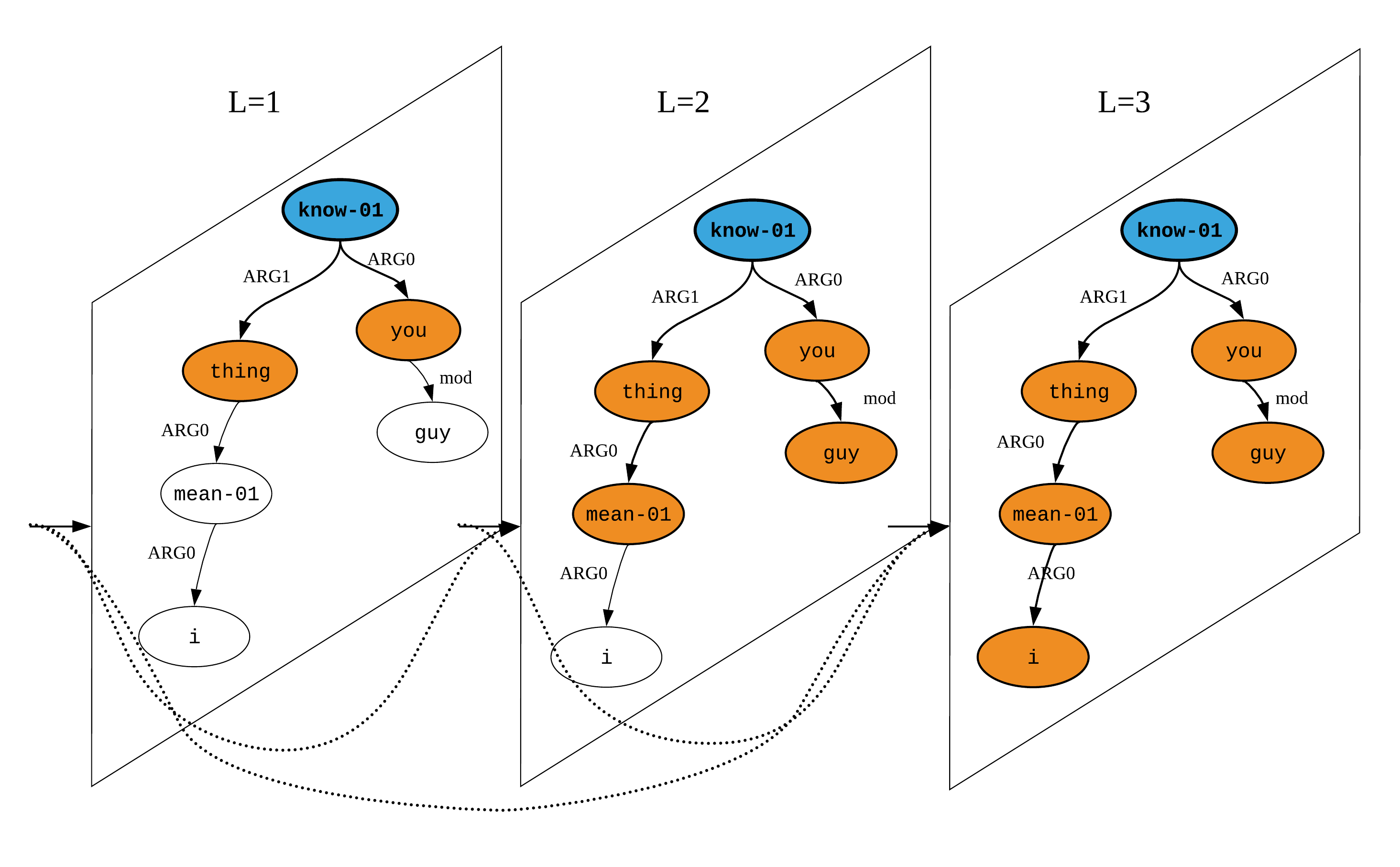}
    \vspace{-7mm}
    \caption{A 3-layer densely connected graph convolutional network. The example AMR graph here corresponds to the sentence ``You guys know what I mean.'' Every layer encodes information about immediate neighbors and 3 layers are needed to capture third-order neighborhood information (nodes that are 3 hops away from the current node). Each layer concatenates all preceding outputs as the input.}
    \vspace{-5mm}
    \label{fig:Figure1}
\end{figure}

Graphs play an important role in natural language processing (NLP) as they are able to capture richer structural information than sequences and trees.
Generally, semantics of sentences can be encoded as graphs. For example, the Abstract Meaning Representation (AMR) \citep{Banarescu2013AbstractMR} is a directed, labeled graph as shown in Figure~\ref{fig:Figure1}, where nodes in the graph denote semantic concepts and edges denote relations between concepts.
Such graph representations can capture rich semantic-level structural information, and are attractive representations useful for semantics related tasks such as semantic parsing \cite{guo2018} and natural language generation \cite{Beck2018GraphtoSequenceLU}.
In this paper, we focus on the graph-to-sequence learning tasks, where we aim at learning representations for graphs which are useful for text generation.

Graph convolutional networks (GCNs) \citep{Kipf2016SemiSupervisedCW} are variants of convolutional neural networks (CNNs) that operate directly on graphs, where the representation of each node is iteratively updated based on those of its adjacent nodes in the graph through an information propagation scheme.
For example, the first layer of GCNs can only capture the graph's adjacency information between immediate neighbors, while with the second layer one will be able to capture second-order proximity information (neighborhood information two hops away from one node) as shown in Figure~\ref{fig:Figure1}.
Formally, $L$ layers will be needed in order to capture neighborhood information that is $L$ hops away. 

GCNs have been successfully applied to many NLP tasks \citep{Bastings2017GraphCE, Zhang2018GraphCO}.
Interestingly, while deeper GCNs with more layers will be able to capture richer neighborhood information of a graph, empirically it has been observed that the best performance is achieved with a 2-layer model \citep{Li2018DeeperII}.
Therefore, recent efforts that leverage recurrence-based graph neural networks have been explored as the alternatives to encode the structural information of graphs.
Examples include graph-state long short-term memory (LSTM) networks  \citep{Song2018AGM} and Gated Graph Neural Networks (GGNNs) \citep{Beck2018GraphtoSequenceLU}. 
Deep architectures based on such recurrence-based models have been successfully built for tasks such as language generation, where rich neighborhood information captured was shown useful.

Compared to recurrent neural networks, convolutional architectures are highly parallelizable and are more amenable to hardware acceleration \citep{Gehring2017ConvolutionalST}.
It is therefore worthwhile to explore the possibility of applying deeper GCNs that are able to capture more non-local information associated with the graph for graph-to-sequence learning.
Prior efforts try to train deep GCNs by incorporating residual connections \citep{Bastings2017GraphCE}.  \citet{Xu2018RepresentationLO} show that vanilla residual connections proposed by \citet{He2016DeepRL} are not effective for graph neural networks. They next attempt to resolve this issue by adding additional recurrent layers on top of graph convolutional layers. However, they are still confined to relatively shallow GCNs architectures (at most 6 layers in their experiments), which may not be able to capture the rich non-local interactions for larger graphs.

In this paper, to better address the issue of learning deeper GCNs, we introduce dense connectivity to GCNs and propose the novel Densely Connected Graph Convolutional Networks (DCGCNs), inspired by DenseNets \citep{Huang2017DenselyCC} that distill insights from residual connections. The dense connectivity strategy is illustrated in Figure~\ref{fig:Figure1} schematically.
Direct connections are introduced from any layer to all its preceding layers. For example, the third layer receives the outputs of the first layer and the second layer, capturing the first-order, the second-order and the third-order neighborhood information.
With the help of dense connections, we are able to train multi-layer GCN models with a large depth, allowing rich local and non-local information to be captured for learning a better graph representation than those learned from the shallower GCN models.

Experiments show that our model is able to achieve better performance for graph-to-sequence learning tasks. For the AMR-to-text generation task, our model surpasses the current state-of-the-art neural models trained on LDC2015E86 and LDC2017T10 by 2 and 4.3 BLEU points respectively. 
For the syntax-based neural machine translation task, our model is also consistently better than others, showing the effectiveness of the model on a large training set. Our code is available at \url{https://github.com/Cartus/DCGCN}.\footnote{Our implementation is based on  MXNET \citep{Chen2015MXNetAF} and the Sockeye \citep{hieber2017sockeye} toolkit.}

\section{Densely Connected GCNs}
\label{sec:3}

In this section, we will present the basic components used for constructing our Densely Connected GCN model. 

\subsection{GCNs}
\label{ssec:3.1}

GCNs are neural networks that operate directly on graph structures \citep{Kipf2016SemiSupervisedCW}.  Here we mathematically illustrate how multi-layer GCNs work on an undirected graph $\mathcal{G} = (\mathcal{V}, \mathcal{E})$, where $\mathcal{V}$ and $\mathcal{E}$ are the set of nodes and edges, respectively. The convolution computation for node $v$ at the $l$-th layer, which takes the input feature representation $\mathbf{h}^{(l-1)}$ as input and outputs the induced representation $\mathbf{h}_v^{(l)}$, can be defined as 
\begin{equation}
\mathbf{h}_{v}^{(l)} = \rho \Big(\sum_{u \in \mathcal{N}(v)} W^{(l)} \mathbf{h}_{u}^{(l-1)} + \mathbf{b}^{(l)} \Big)
\end{equation}
where $W^{(l)}$ is the weight matrix, $\mathbf{b}^{(l)}$ is the bias vector, $\mathcal{N}(v)$ is the set of one-hop neighbors of node $v$, and $\rho$ is an activation function (e.g., RELU \citep{nair2010rectified}). $\mathbf{h}^{(0)}_v$ is the initial input $\mathbf{x}_v$, where $\mathbf{x}_v \in \mathbb{R}^{d}$ and $d$ is the input feature dimension. 

{\bf GCNs with Residual Connections. }
\citet{Bastings2017GraphCE} integrate residual connections~\cite{He2016DeepRL} into GCNs to help information propagation. Specifically, each node is updated according to  Eqn.(1) first and then the resulting representation is combined with the node's representation from the last iteration:
\begin{equation}
\mathbf{h}_{v}^{(l)} = \rho \Big(\sum_{u \in \mathcal{N}(v)} W^{(l)} \mathbf{h}_{u}^{(l-1)} + \mathbf{b}^{(l)} \Big) + \mathbf{h}_{v}^{(l-1)}
\end{equation}

{\bf GCNs with Layer Aggregations. } ~\citet{Xu2018RepresentationLO} propose layer aggregations for GCNs, in which the final representation of each node is computed by combining the node's representations from all GCN layers:
\begin{equation}
\mathbf{h}_{v}^{final} =  LA(\mathbf{h}_{v}^{(l)}, \mathbf{h}_{v}^{(l-1)},...., \mathbf{h}_{v}^{(1)} )
\end{equation}
where the $LA$ function can be concatenation, max-pooling or LSTM-attention operations as defined in~\cite{Xu2018RepresentationLO}.
\subsection{Dense Connectivity}
\label{ssec:3.2}

\begin{figure}[t]
    \centering
    \includegraphics[scale=0.55]{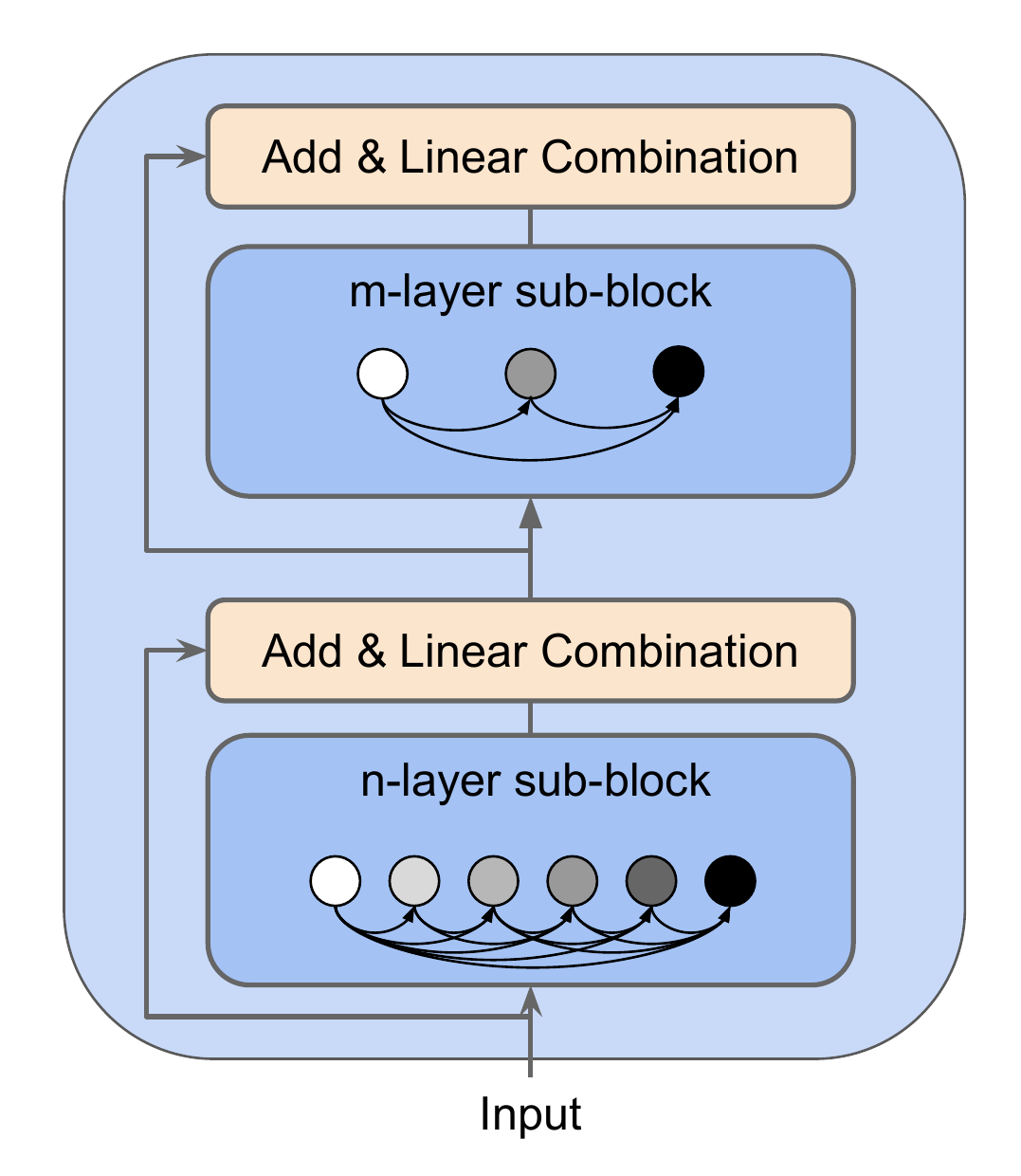}
    \vspace{-6mm}
    \caption{Each DCGCN block has two sub-blocks. Both of them are densely connected graph convolutional layers with different numbers of layers.  A linear transformation is employed between two sub-blocks, followed by a residual connection.}
    \label{fig:Figure2}
    \vspace{-4mm}
\end{figure}

Dense connectivity is the core component of the proposed DCGCN. With dense connectivity, node $v$ in the $l$-th layer not only takes inputs from $\mathbf{h}^{(l-1)}$, but also receives information from all the preceding layers, as shown in Figure~\ref{fig:Figure2}. Mathematically, we first define $\mathbf{g}_{u}^{(l)}$ as the concatenation of the initial node representation and the node representations produced in layers $1$, $\cdots$, $l-1$:
\begin{equation}
\mathbf{g}_{u}^{(l)} = [\mathbf{x}_{u};\mathbf{h}_{u}^{(1)}; ... ;\mathbf{h}_{u}^{(l-1)}]. 
\end{equation}
Such a mechanism allows deeper layers to capture all previous information to alleviate the problem discussed in Section~\ref{sec:1} in graph neural networks. Similar strategies are also proposed in previous works~\cite{He2016DeepRL,Huang2017DenselyCC}. 

\begin{figure}[t]
    \centering
    \includegraphics[scale=0.37]{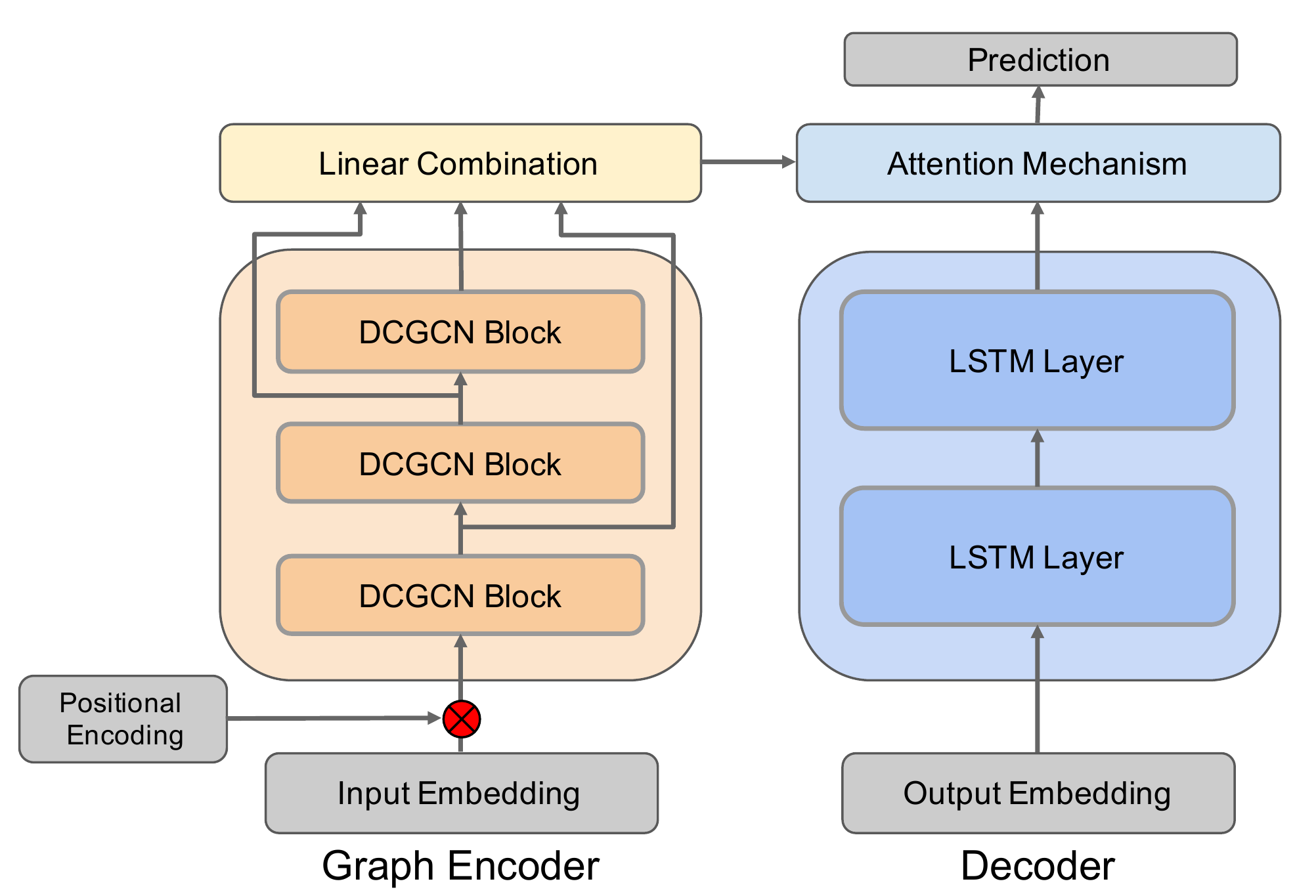}
    \caption{The model concatenates node embeddings and positional embeddings as inputs. The encoder contains a stack of $N$  identical blocks. The linear transformation layer combines output of all blocks into hidden representations. These are fed into an attention mechanism, generating the context vector. The decoder, a 2-layer LSTM \protect \citep{hochreiter1997long}, makes  predictions based on hidden representations and the context vector.}
    \vspace{-3mm}
    \label{fig:Figure3}
\end{figure}

While dense connectivity allows training deeper neural networks, every intermediate layer is designated to be of very small size, allowing adding only a small set of features-maps at each layer. The final classifier makes predictions based on all feature-maps, which is called ``collective knowledge'' \citep{Huang2017DenselyCC}. 
Such a strategy improves the parameter efficiency. In practice, the dimensions of these small hidden layers $d_{hidden}$ are decided by the number of layers $L$ and the input feature dimension $d$. In DCGCN, we use $d_{hidden} = d/L$.

For example, if we have a 3-layer ($L$=3) DCGCN model and input dimension is 300 ($d=300$), the hidden dimension of each layer will be $d_{hidden} = d/L = 300/3=100$. Then we concatenate the output of each layer to form the new representation. We have 3 layers so the output dimension is 300 (3 $\times$ 100). Different from the GCN model whose hidden dimension is larger than or equal to the input dimension, DCGCN model shrinks the hidden dimension as the number of layers increases in order to improve the parameter efficiency similar to DenseNets \citep{Huang2017DenselyCC}.   

Accordingly, we modify the convolution computation of each layer as:
\begin{equation}
\mathbf{h}_{v}^{(l)} = \rho \Big(\sum_{u \in \mathcal{N}(v)} W^{(l)} \mathbf{g}_{u}^{(l)} + \mathbf{b}^{(l)} \Big)
\end{equation}
 The column dimension of the weight matrix increases by $d_{hidden}$ per layer, i.e., $W^{(l)} \in \mathbb{R}^{d_{hidden} \times d^{(l)}}$, where $ d^{(l)} = d+d_{hidden} \times (l-1)$.

\subsection{Graph Attention}
\label{ssec:3.3}

Attention mechanisms have become almost a \textit{de facto} standard in many sequence-based tasks \citep{Vaswani2017AttentionIA}. In DCGCNs, we also incorporate the self-attention strategy by implicitly specifying different weights to different nodes in a neighborhood similar to graph attention networks \citep{Velickovic2017GraphAN}.

In order to perform self-attention on nodes, attention coefficients are required. The input for the calculation is a set of vectors, $\mathbf{\vec{g}}^{(l)} = \{ \mathbf{\vec{g}}_{1}^{(l)}, \mathbf{\vec{g}}_{2}^{(l)}, ..., \mathbf{\vec{g}}_{n}^{(l)} \}$, after node-wise feature transformation $\mathbf{\vec{g}}_{u}^{(l)} = W^{(l)} \mathbf{g}_{u}^{(l)}$. As an initial step, a shared linear projection parameterized by a weight matrix, $W_{a} \in \mathbb{R}^{d_{hidden} \times d_{hidden}}$, is applied to nodes in the graph.  Attention coefficients can be computed as:
\begin{equation}
\alpha_{i j}^{(l)} = \frac{\exp \big( \phi \big( \mathbf{a}^\top[W_{a}\mathbf{\vec{g}}_{i}^{(l)};W_{a}\mathbf{\vec{g}}_{j}^{(l)}]\big) \big)}{\sum_{k \in \mathcal{N}_{i}} \exp \big( \phi \big( \mathbf{a}^\top[W_{a}\mathbf{\vec{g}}_{i}^{(l)};W_{a}\mathbf{\vec{g}}_{k}^{(l)}]\big) \big) }
\end{equation}
where $\mathbf{a} \in \mathbb{R}^{2 d_{hidden}} $ is a weight vector, $\phi$ is the activation function (here we use LeakyReLU \citep{girshick2014rich}). These coefficients are used to compute a linear combination of the node representations. Modifying the convolution computation for attention, we arrive at:
\begin{equation}
\mathbf{h}_{v}^{(l)} = \rho \Big(\sum_{u \in \mathcal{N}(v)} \alpha_{vu}^{(l)} W^{(l)} \mathbf{g}_{u}^{(l)} + \mathbf{b}^{(l)} \Big)
\end{equation}
where $\alpha_{vu}^{(l)}$ are normalized attention coefficients computed by the attention mechanism at $l$-th layer. Note that, these coefficients will not change the dimension of the output representations. 

\section{Graph-to-Sequence Model}
\label{sec:4}

In the following we will explain the model architecture of the graph-to-sequence model. We leverage DCGCNs as the graph encoder, which directly models the graph structure without linearization.

\subsection{Graph Encoder}
\label{ssec:4.1}

The graph encoder is composed of DCGCN blocks, as shown in Figure~\ref{fig:Figure3}.
Within each DCGCN block, we design two types of multi-layer DCGCNs as two sub-blocks to capture graph structure at different abstract levels. As Figure~\ref{fig:Figure2} shows, in each block, the first sub-block has $n$-layers and the second sub-block has $m$-layers. This prototype shares the same spirit with the usage of two different-sized filters in DenseNets \citep{Huang2017DenselyCC}.

{\bf Linear Combination Layer.} In addition to densely connected layers, we include a linear combination layer between multi-layer DCGCNs to filter the representations from different DCGCNs layers, reaching a more expressive representation. This strategy is inspired by  ELMo \citep{Peters2018DeepCW}, which combines the hidden states from different LSTM layers. We also employ a residual connection \citep{He2016DeepRL} to incorporate the initial inputs of multi-layer GCNs into the linear combination layer, see Figure~\ref{fig:Figure3}. Formally, the output of the linear combination layer is defined as:
\begin{equation}
\mathbf{h}_{comb} = W_{comb} \big( \mathbf{h}_{out} +\mathbf{x}_{v} \big) + \mathbf{b}_{comb}
\end{equation}
where $\mathbf{h}_{out}$ is the output of the densely connected layers by concatenating outputs from all previous $L$ layers $\mathbf{h}_{out} = [\mathbf{h}^{(1)}; ... ;\mathbf{h}^{(L)}]$ and $\mathbf{h}_{out} \in \mathbb{R}^{d}$. $\mathbf{x}_{v}$ is the input of the DCGCN layer. $\mathbf{h}_{out}$ and $\mathbf{x}_{v}$ share the same dimension $d$. $W_{comb} \in \mathbb{R}^{d \times d}$ is a weight matrix and $\mathbf{b}_{comb}$ is a bias vector for the linear transformation.  Both $W_{comb}$ and $\mathbf{b}_{comb}$  are different according to different DCGCN layers. In addition,  another linear combination layer is added to get the final representations as shown in Figure~\ref{fig:Figure3}.

\begin{figure}
    \centering
    \includegraphics[scale=0.2]{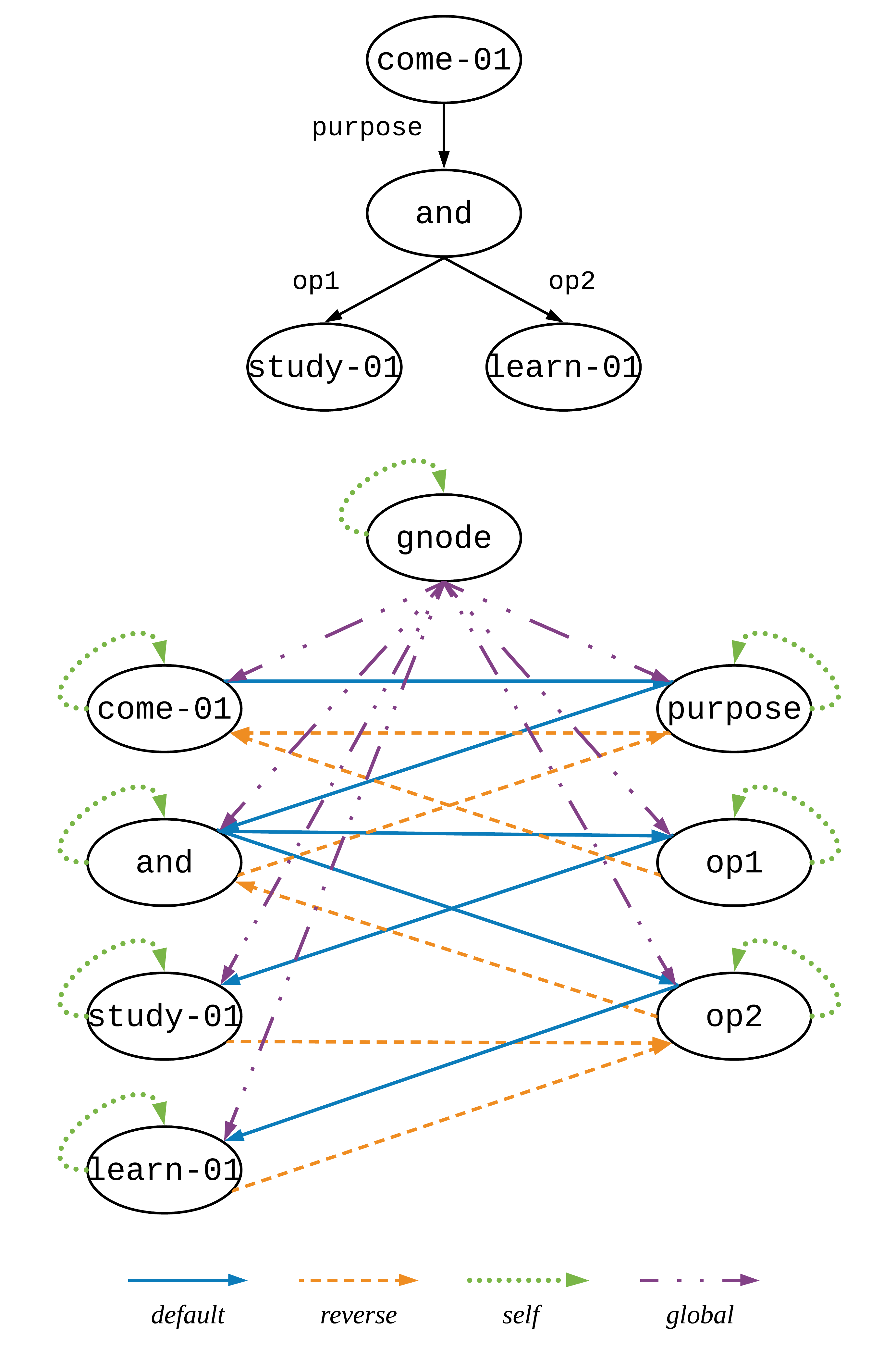}
    \vspace{-2mm}
    \caption{An AMR graph (top) and its corresponding extended Levi graph (bottom). The extended Levi graph contains an additional global node and four different type of edges.}
    \vspace{-6mm}
    \label{fig:Figure4}
\end{figure}

\subsection{Extended Levi Graph}
\label{ssec:4.2}
In order to improve the information propagation process in graph structures such as AMR graphs and dependency trees, previous researchers enrich the original input graphs with additional transformations. \citet{Marcheggiani2017EncodingSW} add \emph{reverse} edges as well as \emph{self-loop} edges  for each node to the original graph. This strategy is similar to the bidirectional recurrent neural networks (RNNs) \citep{elman1990finding}  which can enjoy the information propagation from two directions. \citet{Beck2018GraphtoSequenceLU} adapt this approach and additionally transform the directed input graphs into Levi graphs \citep{Gross:2013:HGT:2613412}. Basically, edges in the original graphs are turned into additional nodes in Levi graphs. With this approach, we can encode the original edge labels and node inputs in the same way. Specifically,  \citet{Beck2018GraphtoSequenceLU} define three types of edge labels on the Levi graph: \emph{default}, \emph{reverse} and \emph{self}, which refer to the original edges, the new virtual edges which are reverse to the original edges and the self-loop edges. 

\citet{Scarselli2009TheGN} add another node that is connected to all other nodes. \citet{Zhang2018SentenceStateLF} uses a global sentence-level node to assemble and back-distribute information. Motivated by these works, we propose \textbf{extended Levi graph}, which adds a global node in Levi graph.  For every node $x$ in the original Levi graph, there is a new edge (\emph{global}) from the global node to $x$. Figure~\ref{fig:Figure4} shows an example AMR graph and its corresponding extended Levi graph.  The edge type vocabulary for the extended Levi graph of the AMR graph now becomes $\mathcal{T} = \{ $\emph{default}, \emph{reverse}, \emph{self}, \emph{global}$\}$.  Our motivations are three-folds. First, the global node gives each node a global view of the input graph, which can make each node more aware of the non-local information.  Second, the global node can  serve as a hub to help node communications, which can facilitate the node information propagation process. Third, the output vectors of the global node in the encoder can be used as the initial states of the decoder, which are crucial for sequence-to-sequence learning tasks. Prior efforts average
representations of all nodes as the graph embedding to initialize the decoder. Instead, we directly use the learned representation of the global nodes, which captures the information from all nodes in the whole graph.

\begin{figure}
    \centering
    \includegraphics[scale=0.28]{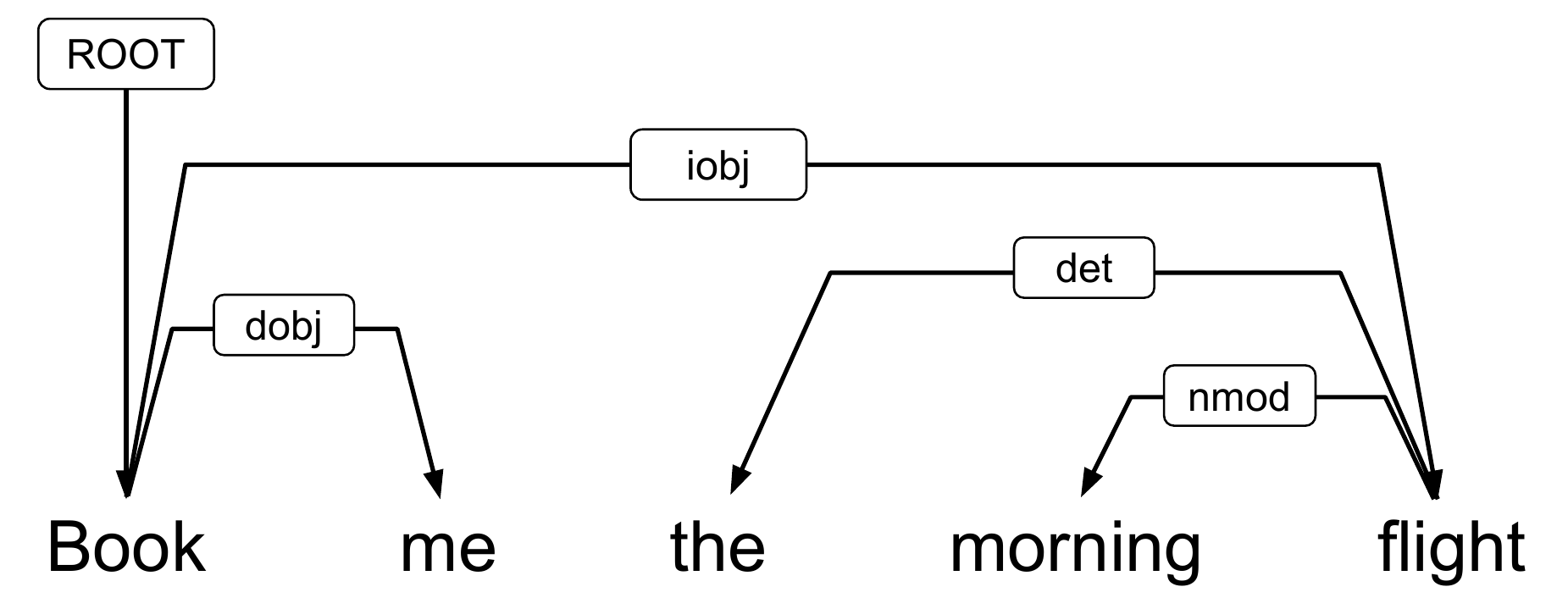}
    \includegraphics[scale=0.27]{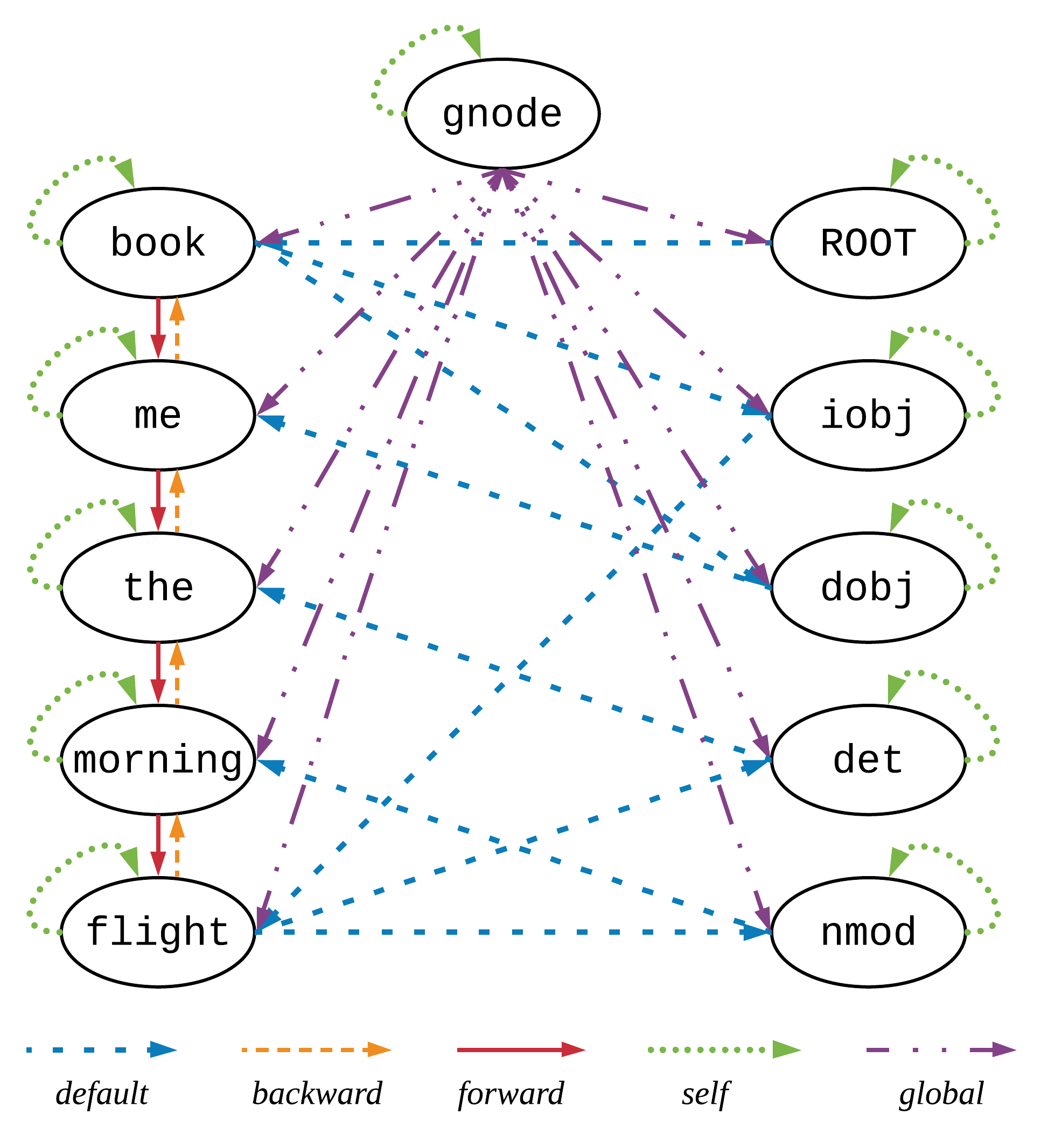}
    \vspace{-2mm}
    \caption{A dependency tree and its extended Levi graph.}
    \vspace{-5mm}
    \label{fig:Figure5}
\end{figure}

The input to the syntax-based neural machine translation task is the dependency tree. Unlike the AMR graph, the sentence contains significant sequential information. \citet{Beck2018GraphtoSequenceLU} inject this information by adding sequential connections to each token. In our model, we also add forward and backward sequential connections as illustrated in Figure~\ref{fig:Figure5}. Therefore, the edge type vocabulary for the extended Levi graph of the dependency tree becomes $\mathcal{T} = \{$\emph{default}, \emph{reverse}, \emph{self}, \emph{global}, \emph{forward}, \emph{backward}$\}$.

Positional encodings about the relative or absolute position of the tokens have been proved beneficial for sequence learning \citep{Gehring2017ConvolutionalST}. We also include positional encodings by concatenating them with the learned word embeddings. The positional encodings are indexed by integer values representing the minimum distance from the root node. For example, $\mathsf{come}$-$\mathsf01$ in Figure~\ref{fig:Figure4} is the root node of the AMR graph, so its index should be 0, where $\mathsf{and}$ is the child node of $\mathsf{come}$-$\mathsf01$, its index is 1. Notice that we denote the index of the global node as -1.

\subsection{Direction Aggregation}
\label{ssec:4.3}

Directionality and edge labels play an important role in linguistic structures. Information from incoming edges, outgoing edges and self edges should be treated differently by using separate weight matrices. Moreover, information from incoming edges that have different labels should have different weight matrices too. Following this motivation, we incorporate the directionality of an edge directly in its label. For example, node $\mathsf{learn}$-$\mathsf{01}$ in Figure~\ref{fig:Figure4} has three incoming edges, these edges have three different types: $default$ (from node $\mathsf{op2}$), $self$ (from node $\mathsf{learn}$-$\mathsf{01}$) and $global$ (from node $\mathsf{gnode}$). For AMR graph we have four types of edges while for dependency trees we have six as mentioned in Section ~\ref{ssec:4.2}. Thus, considering different type of edges, we modify the convolution computation as:
\begin{equation}
\begin{aligned}
\mathbf{v}_{t}^{(l)} = \rho \Big(\sum_{ \substack {u \in \mathcal{N}(v)\\dir(u, v)=t } }\alpha_{vu}^{(l)} W_{t}^{(l)} \mathbf{g}_{u}^{(l)} + \mathbf{b}_{t}^{(l)} \Big)
\end{aligned}
\end{equation}
where $dir(u, v)$ selects the weight matrix and bias term associated with the edge type $t$. For example, in the AMR generation task, there are four edge types: \emph{default}, \emph{reverse}, \emph{self} and \emph{global}. Each type corresponds to a separate weight matrix and a separate bias term.

Now we need to aggregate representations learned from different types of edges. A simple way to do this is averaging them to get the final representations. However, \citet{Hamilton2017InductiveRL} show that using a mean-based function to aggregate feature information from different nodes may not be satisfactory, since information from different sources should not be treated equally. Thus we assign different weights to information from different types of edges to integrate such information. Specifically, we concatenate the learned representations from all types of edges and perform a linear transformation,  mathematically illustrated as:
\begin{equation}
    f([\mathbf{v}_{1}^{(l)};\cdots;\mathbf{v}_{T}^{(l)}]) = W_{f} [\mathbf{v}_{1}^{(l)};\cdots;\mathbf{v}_{T}^{(l)}] + \mathbf{b}_{f} 
\end{equation}
where $W_{f} \in \mathbb{R}^{d^{'} \times d_{hidden}}$ is the weight matrix and $d^{'} = T \times d_{hidden}$. $T$ is the size of the edge type vocabulary and $d_{hidden}$ is the hidden dimension  in DCGCN layers as described in Section ~\ref{ssec:3.2}.
$\mathbf{b}_{f} \in \mathbb{R}^{d_{hidden}} $ is a bias vector. Finally, the convolution computation becomes:
\begin{equation}
\mathbf{h}_{v}^{(l)} = \rho \Big( f([\mathbf{v}_{1}^{(l)};\cdots;\mathbf{v}_{T}^{(l)}]) \Big)
\end{equation}
\subsection{Decoder}

We use an attention-based LSTM decoder \cite{Bahdanau2014NeuralMT}. The initial state of the decoder is the representation of the global node described in Section 3.2. The decoder yields the natural language sequence by calculating a sequence of hidden states sequentially. Here we also include the coverage mechanism \cite{Tu2016ModelingCF}. Therefore, when generating the $t$-th token, the decoder considers five factors: the attention memory, the word embedding of the $(t-1)$-th token, the previous hidden state of LSTM, the previous context vector and the previous coverage vector.

\section{Experiments}

\input{tables/table1stat.tex}

\subsection{Experimental Setup}

We assess the effectiveness of our models on two typical graph-to-sequence learning tasks, including AMR-to-text generation and syntax-based neural machine translation (NMT).  For the AMR-to-text generation task, we use two benchmarks --- the LDC2015E86 dataset (AMR15) and the LDC2017T10 dataset (AMR17). In these datasets, each instance contains a sentence and an AMR graph. We follow \citet{Konstas2017NeuralAS} to apply entity simplification in the preprocessing steps.  We then transform each  preprocessed AMR graph into its extended Levi graph as described in Section~\ref{ssec:4.2}. For the syntax-based NMT task, we evaluate our model  on both the En-De and the En-Cs News Commentary v11 dataset from the WMT16 translation task\footnote{\scriptsize{\url{http://www.statmt.org/wmt16/translation-task.html}}}. We parse English sentences after tokenization to generate the dependency trees on the source side  using SyntaxNet \citep{Alberti2017SyntaxNetMF}\footnote{\scriptsize{\url{https://github.com/tensorflow/models/tree/master/research/syntaxnet}}}. We tokenzie Czech and German using the Moses tokenizer\footnote{\scriptsize{\url{https://github.com/moses-smt/mosesdecoder}}}. 
On the target side, we use byte-pair encodings (BPE) \citep{Sennrich2016NeuralMT} with 8,000 merge operations to obtain subwords. We transform the labelled dependency trees into their corresponding extended Levi graphs as described in Section ~\ref{ssec:4.2}. Table~\ref{tab:stat} shows the statistics of these four datasets. The AMR-to-text datasets contain about 16K $\sim$ 36K training instances. The NMT datasets are relatively large, consisting of around 200K training instances. 

We tune model hyper-parameters using random layouts based on the results of the development set. We choose the number of DCGCN blocks ($Block$) from $\{1,2,3, 4\}$. We select the feature dimension $d$ from $\{180, 240, 300, 360, 420\}$. We do not use pretrained embeddings. The encoder and the decoder share the training vocabulary. We adopt Adam \citep{Kingma2014AdamAM} with an initial learning rate 0.0003  as the optimizer. The batch size ($Batch$) candidates are $\{16, 20, 24\}$. We determine when to stop training based on the perplexity change in the development set. For decoding, we use beam search with beam size 10. Through preliminary experiments, we find that the combinations ($Block$=$4$, $d$=$360$, $Batch$=$16$) and ($Block$=$2$,  $d$=$360$, $Batch$=$24$) give best results on AMR and NMT tasks, respectively. Following previous works, we evaluate the results in terms of both BLEU (\texttt{B}) scores \citep{Papineni2002BleuAM} and sentence-level CHRF++ (\texttt{C}) scores  \citep{Popovic2017chrFWH,Beck2018GraphtoSequenceLU}. Particularly, we use case insensitive BLEU scores for AMR and case sensitive BLEU scores for NMT.  For ensemble models, we train five models  with different random seeds and then use Sockeye \citep{hieber2017sockeye} to perform default ensemble decoding. 

\subsection{Main Results on AMR-to-text Generation}

\input{tables/table5amrmain.tex}

We compare the performance of DCGCNs with the other three kinds of models: (1) sequence-to-sequence (Seq2Seq) models which use linearized graphs as inputs; (2) graph encoders (\text{GGNN2Seq}, \text{GraphLSTM}, \text{GCNSEQ}); (3) models trained with external resources. For convenience, we denote the LSTM-based Seq2Seq models of \citet{Konstas2017NeuralAS} and \citet{Beck2018GraphtoSequenceLU} as \text{Seq2SeqK}  and \text{Seq2SeqB}, respectively. GGNN2Seq \citep{Beck2018GraphtoSequenceLU} is the model that leverages GGNNs as graph encoders. GCNSEQ \citep{Damonte2019StructuralNE} applies a bidirectional RNN on top of the 2-layer GCNs as the encoder. 

Table~\ref{tab:mainamr} shows the results on AMR17. Our single model achieves 27.6 BLEU points, which is the new state-of-the-art result for single models. In particular, our single \text{DCGCN} model consistently outperforms Seq2Seq models by a significant margin when trained without external resources. For example, the single \text{DCGCN} model gains 5.9 more BLEU points than the single models of \text{Seq2SeqB} on AMR17. These results demonstrate the importance of explicitly capturing the graph structure in the encoder.

In addition, our single \text{DCGCN} model obtains better results than previous ensemble models. For example, on AMR17, the single \text{DCGCN} model is 1 BLEU point higher than the ensemble model of \text{Seq2SeqB}. Our model requires substantially fewer parameters, e.g., the parameter size is only 3/5 and 1/9 of those in \text{GGNN2Seq} and \text{Seq2SeqB}, respectively. The ensemble approach based on combining five DCGCN models initialized with different random seeds achieves a BLEU score of 30.4 and a CHRF++ score of 59.6.

Under the same setting, our model also consistently outperforms graph encoders based on recurrent neural networks or gating mechanisms. For \text{GGNN2Seq}, our single model is 3.3 and 0.1 BLEU points higher than their single and ensemble models, respectively. We also have similar observations in term of CHRF++ scores for sentence-level evaluations. \text{DCGCN} also outperforms \text{GraphLSTM} by 2.0 BLEU points in the fully supervised setting as shown in Table ~\ref{tab:mainamrexternal}. Note that \text{GraphLSTM} uses char-level neural representations and pretrained word embeddings, while our model solely relies on word-level representations with random initializations. 
This empirically shows that compared to recurrent graph encoders, DCGCNs can learn better representations for graphs. 
For \text{GCNSEQ}, our single models are 3.1 and 1.3 BLEU points higher than their models trained on AMR17 and AMR15 dataset, respectively. These results demonstrate that DCGCNs are able to capture contextual information without relying on additional RNNs.

\input{tables/table6amrexternal.tex}
\input{tables/table7nmtmain.tex}

Moreover, we compare our results with the state-of-the-art semi-supervised models on the AMR15 test set (Table ~\ref{tab:mainamrexternal}), including  non-neural methods such as TSP \citep{Song2016AMRtotextGA}, PBMT \citep{Pourdamghani2016GeneratingEF}, Tree2Str \citep{Flanigan2016GenerationFA} and SNRG \citep{Song2017AMRtotextGW}. All these non-neural models train language models on the whole Gigaword corpus. Our ensemble model gives 28.2 BLEU points without external data, which is  better than them.

Following \citet{Konstas2017NeuralAS, Song2018AGM}, we also evaluate our model using external Gigaword sentences as training data. We first use the additional data to pretrain the model, then fine-tune it on the gold data. Using additional 0.1M data, the single DCGCN model achieves a BLEU score of 29.0, which is higher than Seq2SeqK \citep{Konstas2017NeuralAS} and GraphLSTM \citep{Song2018AGM} trained with 0.2M additional data. When using the same amount of 0.2M data, the performance of DCGCN is 4.2 and 3.4 BLEU points higher than Seq2SeqK and GraphLSTM. DCGCN model is able to achieve a competitive BLEU points (33.2) by using 0.3M external data, while GraphLSTM achieves a score of 33.6 by using 2M data and Seq2SeqK achieves a score of 33.8 by using 20M data. These results show that our model is more effective in terms of using automatically generated AMR graphs. Using 0.3M additional data, our ensemble model achieves the new state-of-the-art result of 35.3 BLEU points. 

\subsection{Main Results on Syntax-based NMT} 

Table ~\ref{table3} shows the results for the English-German (En-De) and English-Czech (En-Cs) translation tasks. \text{BoW+GCN}, \text{CNN+GCN} and  \text{BiRNN+GCN} refer to employing the following encoders with a GCN layer on top respectively: 1) a bag-of-words encoder, 2) a one-layer CNN, 3) a bidirectional RNN. \text{PB-SMT} is the phrase-based statistical machine translation model using Moses \citep{Koehn2007MosesOS}.  
Our single model achieves 19.0 and 12.1 BLEU points on the En-De and En-Cs tasks, respectively, significantly outperforming all the single models. For example, compared to the best GCN-based model  (\text{BiRNN+GCN}), our single \text{DCGCN} model surpasses it by 2.7 and 2.5 BLEU points on the En-De and En-Cs tasks, respectively. Our models consist of full GCN layers, removing the burden of employing a recurrent encoder to extract non-local contextual information in the bottom layers.  Compared to non-GCN models,  our single \text{DCGCN} model is 2.2 and 1.9 BLEU points higher than the current state-of-the-art single model (\text{GGNN2Seq}) on the En-De and En-Cs translation tasks, respectively. In addition,  our single model is comparable to the ensemble results of \text{Seq2SeqB} and \text{GGNN2Seq}, while the number of parameters of our models is only about 1/6 of theirs. Additionally, the  ensemble DCGCN models achieve 20.5 and 13.1 BLEU points on the En-De and En-Cs tasks, respectively.  Our ensemble results are significantly higher than those of the state-of-the-art syntax-based ensemble models reported by \text{GGNN2Seq} (En-De: 20.5 v.s. 19.6; En-Cs: 13.1 v.s. 11.7 in terms of BLEU).  

\subsection{Additional Experiments}

\paragraph{Layers in the sub-block.}

Table \ref{tab:devnm} shows the effect of the number of layers of each sub-block on the AMR15 development set. DenseNets \citep{Huang2017DenselyCC} use two kinds of convolution filters: $1\times1$ and $3\times3$. Similar to DenseNets, we choose the values of $n$ and $m$ for layers from $[1,2,3,6]$. We choose this value range by considering the scale of non-local nodes, the abstract information at different level and the calculation efficiency. For brevity, we only show representative configurations. We first investigate DCGCN with one block.  In general, the performance increases when we gradually enlarge  $n$ and $m$. For example, when $n$=$1$ and $m$=$1$, the BLEU score is 17.6; when $n$=$6$ and $m$=$6$, the BLEU score becomes 22.0. We observe that the three settings ($n$=$6$, $m$=$3$), ($n$=$3$, $m$=$6$) and ($n$=$6$, $m$=$6$) give similar results for both 1 DCGCN block and 2 DCGCN blocks. Since the first two settings contain less parameters than the third setting, it is reasonable  to choose either ($n$=$6$, $m$=$3$) or ($n$=$3$, $m$=$6$). For later experiments, we use ($n$=$6$, $m$=$3$). 

\input{tables/table2devnum.tex}
\input{tables/table3baseline.tex}

\paragraph{Comparisons with Baselines.} The first block in Table \ref{tab:baselines} shows the performance of our two baseline models: multi-layer GCNs with residual connections (\text{GCN+RC}) and multi-layer GCNs with both  residual connections and layer aggregations (\text{GCN+RC+LA}). In general, increasing the number of GCN layers from 2 to 9 boosts the model performance. However, when the layer number exceeds 10, the performance of both baseline models start to drop. For example, \text{GCN+RC+LA} (10) achieves a BLEU score of 21.2, which is worse than \text{GCN+RC+LA} (9). In preliminary experiments, we cannot manage to train very deep \text{GCN+RC} and \text{GCN+RC+LA}  models. In contrast, our \text{DCGCN} models can be trained using a large number of layers. For example, \text{DCGCN4} contains 36 layers. When we increase the \text{DCGCN} blocks from 1 to 4, the model performance continues increasing on AMR15 development set. We therefore choose \text{DCGCN4} for the AMR experiments. Using a similar method, \text{DCGCN2} is selected for the NMT tasks. When the layer numbers are 9, \text{DCGCN1} is better than \text{GCN+RC}  in term of \texttt{B}/\texttt{C} scores (21.7/51.5 v.s. 21.1/50.5). \text{GCN+RC+LA} (9) is sightly better than \text{DCGCN1}. However, when we set the number to 18, \text{GCN+RC+LA} achieves a BLEU score of 19.4, which is significantly worse than the BLEU score obtained by \text{DCGCN2} (23.3). We also try \text{GCN+RC+LA} (27), but it does not converge. In conclusion, these results above can show the robustness and effectiveness of our DCGCN models.

\input{tables/table2dcgcnparams.tex}

\begin{figure*}
    \centering
    \includegraphics[scale=0.52]{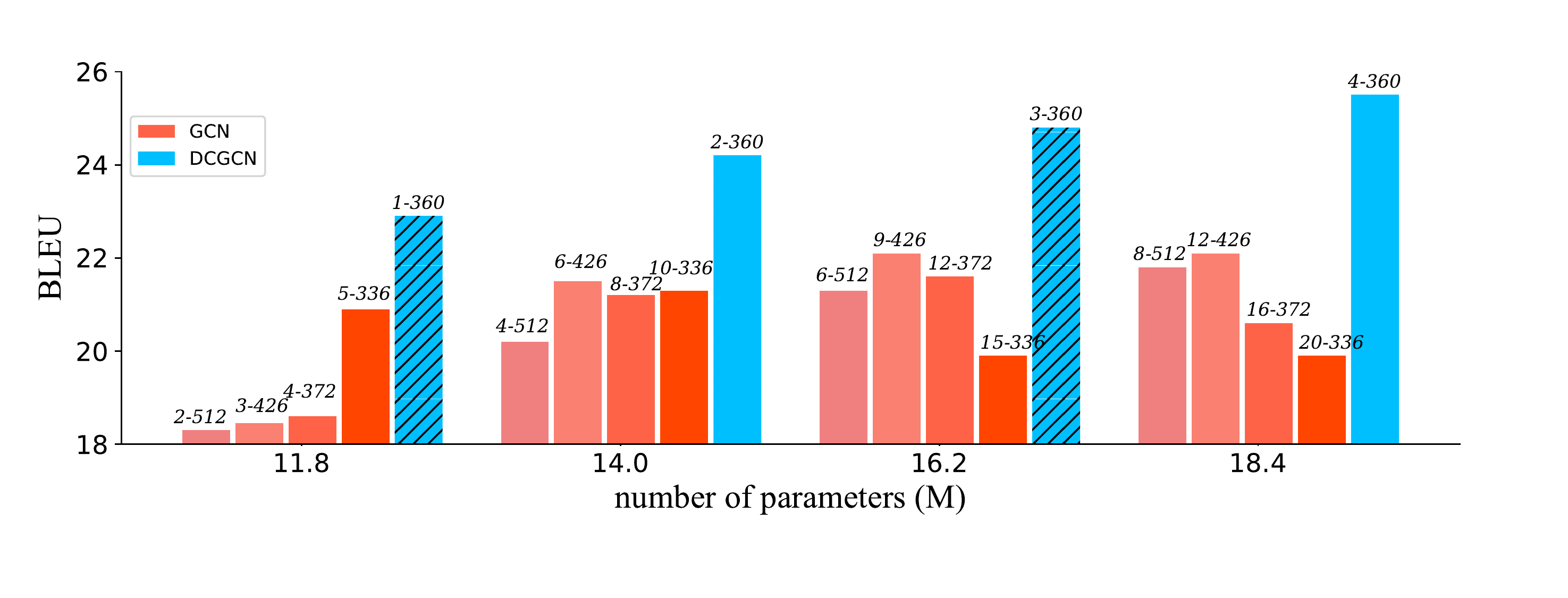}
    \vspace{-2.8em}
    \caption{Comparison of DCGCN and GCN over different number of parameters. $a$-$b$ means the model has $a$ layers ($a$ blocks for DCGCN) and the hidden size is $b$ (e.g., 5-336 means a 5-layers GCN with the hidden size 336).}
    \label{fig:Figure7}
    \vspace{-1.2em}
\end{figure*}

\paragraph{Performance v.s. Parameter Budget.} We also evaluate the performance of DCGCN model against different number of parameters on the AMR generation task. Results are shown in Figure~\ref{fig:Figure7}. Specifically, we try four parameter budgets, including 11.8M, 14.0M, 16.2M and 18.4M. These numbers correspond to the model size (in terms of number of parameters) of DCGCN1, DCGCN2, DCGCN3 and DCGCN4, respectively. For each budget, we vary both the depth of GCN models and the hidden vector dimensions of each node in GCNs in order to exhaust the entire budget. For example, $GCN(2)-512$, $GCN(3)-426$, $GCN(4)-372$ and $GCN(5)-336$ contain about 11.8M parameters, where $GCN(i)-d$ indicates a GCN model with $i$ layers and the hidden size for each node is $d$. We compare DCGCN1 with these four models. DCGCN1 gives 22.9 BLEU points. For the GCN models, the best result is obtained by $GCN(5)-336$, which falls behind DCGCN1 by 2.0 BLEU points. We compare DCGCN2, DCGCN3 and DCGCN4 with their equal-sized GCN models in a similar way. The results show that DCGCN consistently outperforms GCN under the same parameter budget. When the parameter budget becomes larger, we can observe that the performance difference becomes more prominent. In particular, the BLEU margins between DCGCN models and their best GCN models are 2.0, 2.7, 2.7 and 3.4, respectively.

\paragraph{Performance v.s. Layers.}  We compare DCGCN models with different layers under the same parameter budget. Table \ref{tab:gcnparams} shows the results. For example, when both DCGCN1 and DCGCN2 are limited to 10.9M parameters, DCGCN2 obtains 22.2 BLEU points, which is higher than DCGCN1 (20.9). Similarly, when DCGCN3 and DCGCN4 contain 18.6M and 18.4M parameters. DCGCN4 outperforms DCGCN3 by 1 BLEU point with a slightly smaller model. In general, we found when the parameter budget is the same, deeper DCGCN models can obtain better results than the shallower ones. 

\input{tables/table4density.tex}
\paragraph{Level of Density.} Table \ref{tab:density} shows the ablation study of the level of density of our model.  We use DCGCNs with 4 dense blocks as the full model. Then we remove dense connections gradually from the last block to the first block.  In general, the performance of the model drops substantially as we remove more dense connections until it cannot converge without dense connections. The full model gives 25.5 BLEU points on the AMR15 dev set. After removing the dense connections in the last block, the BLEU score becomes 24.8. Without using the dense connections in the last two blocks, the score drops to 23.8.  Furthermore, excluding the dense connections in the last three blocks only gives 23.2 BLEU points. Although these four models have the same number of layers, dense connections allow the
model to achieve much better performance. If all the dense connections are not considered, the model does not coverage at all. These results indicate dense connections do play a significant role in our model.

\input{tables/table4ablation.tex}

\paragraph{Ablation Study for Encoder and Decoder.} 
Following \citet{Song2018AGM}, we conduct a further ablation study for modules used in the graph encoder and LSTM decoder on the AMR15 dev set, including linear combination, global node, direction aggregation, graph attention mechanism and coverage mechanism using the 4-block models by always keeping the dense connections.

Table~\ref{tab:devablation} shows the results. For the encoder, we find that the linear combination and the global node have more contributions in terms of \texttt{B}/\texttt{C} scores. The results drop by 2/2.2 and 1.3/1.2 points respectively after removing them. Without these two components, our model gives a BLEU score of 22.6, which is still better than the best GCN+RC model (21.1) and the best GCN+RC+LA model (22.1). Adding either the global node or the linear combination improves the baseline models with only dense connections. This suggests that enriching input graphs with the global node and including the linear combination can facilitate GCNs to learn better information aggregations, producing more expressive graph representations. Results also show the linear combination is more effective than the global node. Considering them together further enhances the model performance.  After removing the graph attention module, our model gives 24.9 BLEU points.  Similarly, excluding the direction aggregation module leads to a performance drop to 24.6 BLEU points. The coverage mechanism is also effective in our models. Without the coverage mechanism, the result drops by 1.7/2.4 points for \texttt{B}/\texttt{C} scores.

\subsection{Analysis and Discussion}

\begin{figure}[!t]
    \centering
    \includegraphics[scale=0.4]{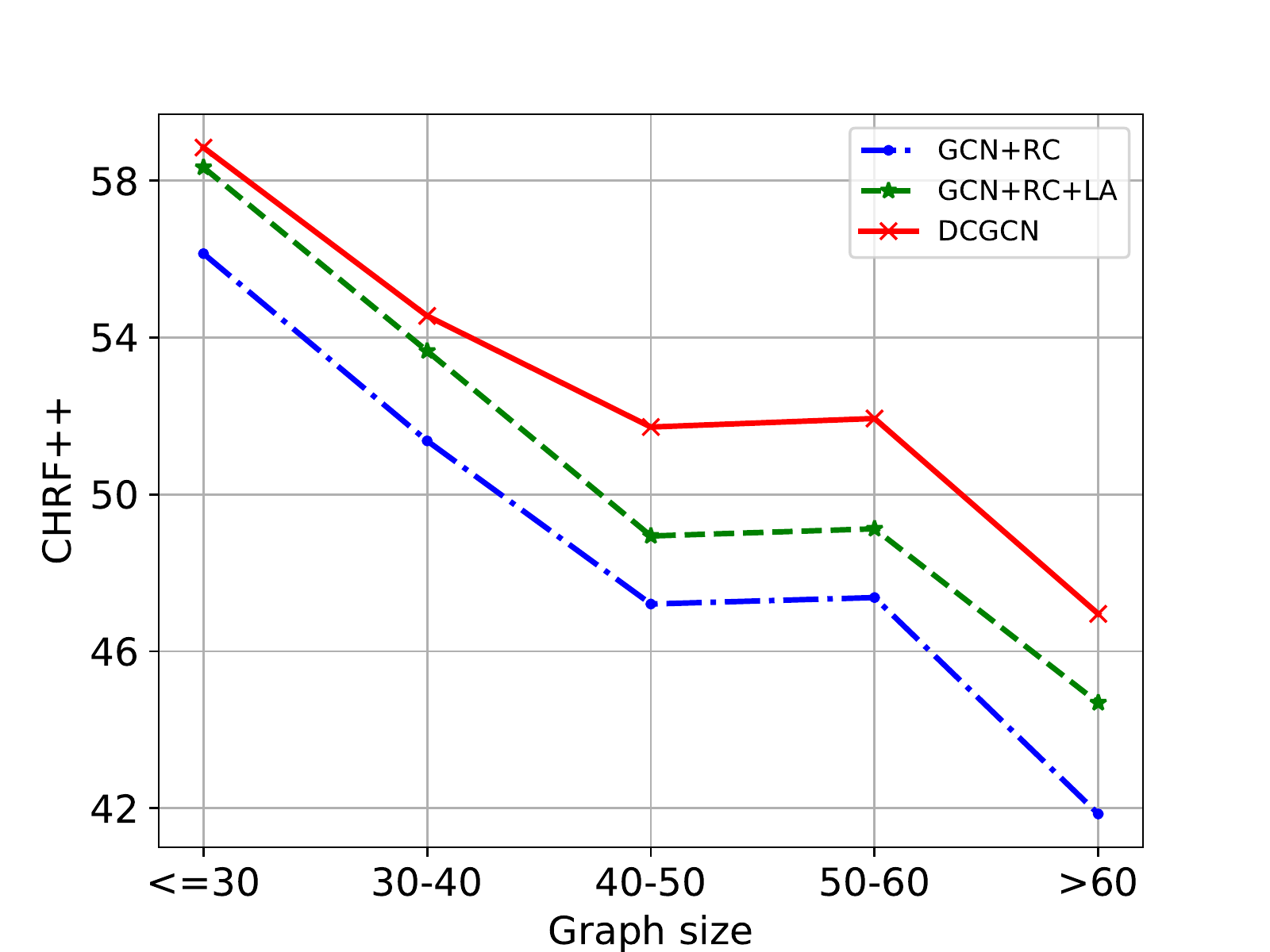}
    \vspace{-2mm}
    \caption{CHRF++ scores with respect to the input graph size for three models.}
    \label{fig:Figure6}
    \vspace{-4mm}
\end{figure}

\paragraph{Graph size.} Following~\citet{Bastings2017GraphCE}, we show in Figure~\ref{fig:Figure6} the CHRF++ score variations according to the graph size $|G|$ on the AMR2015 development set, where $|G|$ refers to the number of nodes in the extended Levi graph. We bin the graph size into five classes $(\leq 30, (30,40], (40,50], (50,60], >60)$. We average the sentence-level CHRF++ scores of the sentences in the same bin to plot Figure~\ref{fig:Figure6}.  For small graphs (i.e., $|G|\leq  30$), DCGCN obtains similar results as the baselines. For large graphs, DCGCN significantly outperforms the two baselines. In general, as the graph size increases, the gap between DCGCN and the two baselines becomes larger.  In addition, we can also notice that the margin between GCN and GCN+LA is quite stable, while the margin between DCGCN and GCN+LA varies according to the graph size. The trend for BLEU scores is similar to CHRF++ scores. This suggests that DCGCN can perform better for larger graphs as its deeper architecture can capture the long-distance dependencies. Dense connections facilitate  information propagation in large graphs, while shallow GCNs might struggle to capture such dependencies.

\paragraph{Example output.} 
\input{tables/table8output.tex}
Table \ref{tab:output} shows example outputs from three models for the AMR-to-text task, together with the corresponding AMR graph as well as the text reference.   The word ``technology'' in the reference acts as a link between ``global trade'' and ``weapons of mass destruction'', offering the background knowledge to help understand the context. The word ``instructions'' also plays a crucial role in the generated sentence -- without the word the sentence will have a significantly different meaning. Both GCN+RC and GCN+RC+LA fail to successfully generate these two important words. The output from GCN+RC does not even appear to be grammatically correct. In contrast, DCGCN manages to generate both words. We believe this is because DCGCN is able to learn richer semantic information by capturing complex long dependencies. GCN+RC+LA does generate an output which looks similar to the reference at the token level. However, the conveyed semantic information in the generated sentence largely differs from that of the reference. DCGCNs do not have this problem. 

\section{Related Work}
\label{sec:2}

Our work builds on a rich line of recent efforts on graph-to-sequence models, graph convolutional networks and densely connected convolutional networks.


\paragraph{Graph-to-sequence learning.} 
Early research efforts for graph-to-sequence learning are based on statistical methods. \citet{Lu2009NaturalLG} present a language generation model using the tree-structured meaning representation based on tree conditional random fields. \citet{Lu2011APF} propose a model for language generation from lambda calculus expressions which can be represented as forest structures. \citet{Konstas2012UnsupervisedCG, Konstas2013InducingDP} leverage hypergraphs for concept-to-text generation. \citet{Flanigan2016GenerationFA} transform a given AMR graph into a spanning tree, before translating it into a sentence using a tree-to-string transducer. \citet{Pourdamghani2016GeneratingEF} adopt a phrase-based model for machine translation \citep{Koehn2003StatisticalPT} based on a linearized AMR graph. \citet{Song2017AMRtotextGW} leverage a synchronous node replacement grammar. {\color{black}\citet{Konstas2017NeuralAS} also linearize the input graph and feed it to the Seq2Seq model \cite{Sutskever2014SequenceTS}}. 

Sequence based neural networks may lose structural information from the original graph since they require linearization of the input graph. Recent research efforts consider developing encoders with graph neural networks. \citet{Beck2018GraphtoSequenceLU} employ GGNNs \citep{Li2015GatedGS} as the encoder and introduce the Levi graph that allows nodes and edges to have their own hidden representations. \citet{Song2018AGM} propose the graph-state LSTM to directly encode graph-level semantics. In order to capture non-local information, the encoder performs graph state transition by information exchange between connected nodes. Their work belongs to the family of recurrent neural networks (RNN). Our graph encoder is built based on the graph convolutional networks (GCNs). Recurrent graph neural networks \citep{Li2015GatedGS, Song2018AGM} use gated operations to update node states while graph convolutional networks use linear transformation. The contrast between our model and theirs is reminiscent of the contrast between CNN and RNN. 

Closest to our work, \citet{Bastings2017GraphCE, Damonte2019StructuralNE} stack GCNs upon a RNN or CNN encoder since 2-layer GCNs may not be able to capture non-local information, especially when the graph is large. \citet{marcheggiani-perez-beltrachini-2018-deep} also leverages purely GCN model for encoding the graph structure, but their model is still confined to relatively shallow architecture. Our graph encoder solely relies on the DCGCN model, whose deep network structure encodes richer local and non-local information for learning better graph representations.

\paragraph{Densely connected convolutional networks.}
Intuitively, neural networks should be able to learn rich representations by stacking a large number of layers.
However, empirical results often do not support such an intuition -- useful information captured in earlier layers may get lost after passing through subsequent layers.
Many recent efforts focus on resolving such an issue.
Highway Networks \citep{Srivastava2015TrainingVD} use bypassing paths along with gating units to train networks.
ResNets \citep{He2016DeepRL}, in which identity mappings are used as bypassing paths, have achieved impressive performance on various tasks.
DenseNets \citep{Huang2017DenselyCC} refine this insight and propose a dense connectivity strategy, which connects all layers directly with each other to ensure maximum information flow between layers. 


\paragraph{Graph convolutional networks.} Early efforts that attempt to extend neural networks to deal with arbitrary structured graphs are introduced by \citet{Gori2005ANM} and \citet{Scarselli2009TheGN}, where
the states of nodes are updated based on the states of their neighbors.
{\color{black}\citet{Bruna2014SpectralNA} then applies the convolution operation on graph Laplacians to construct efficient architectures in the spectral domain. Subsequent efforts improve its computational efficiency with local spectral convolution techniques \citep{Henaff2015DeepCN,Defferrard2016ConvolutionalNN, Kipf2016SemiSupervisedCW}.}

Our approach is closely related to GCNs \citep{Kipf2016SemiSupervisedCW}, {\color{black}which restrict the filters to operate on a first-order neighborhood around each node.}  Recent improvements and extensions of GCNs include using additional aggregation methods such as vertex attention \citep{Velickovic2017GraphAN} or pooling mechanism \citep{Hamilton2017InductiveRL} to better summarize neighborhood states. 

However, the best performance of GCNs is achieved with a 2-layer model  while deeper models perform worse though they can potentially have access to more non-local information. \citet{Li2018DeeperII} shows that this issue is due to the over-smoothed output representations that impede distinguishing nodes from different clusters. Recent attempts that try to address this issue includes the use of layer-aggregation functions \citep{Xu2018RepresentationLO}, which combine learned features from all layers, and the use of co-training and self-training mechanisms that encourage exploration on the entire graph  \citep{Li2018DeeperII}.

\section{Conclusion}

We introduce the novel densely connected graph convolutional networks (DCGCNs) to learn structural graph representations.
Experimental results show that DCGCNs can outperform state-of-the-art models in two tasks: AMR-to-text generation and syntax-based neural machine translation. Unlike previous designs of GCNs, DCGCNs scale naturally to significantly more layers without suffering from performance degradation and optimization difficulties, thanks to the introduced dense connectivity mechanism.
Such a deep architecture allows the encoder to better capture the rich structural information of a graph, especially when it is large.

There are multiple venues for future work. One natural question we would like to ask is how to make use of the proposed framework to perform improved graph representation learning for various graph related tasks \cite{Xu2018RepresentationLO}.
On the other hand, we would also like to investigate how other NLP applications such as relation extraction \citep{Zhang2018GraphCO} and semantic role labeling \citep{Marcheggiani2017EncodingSW} can potentially benefit from our proposed approach.

\section*{Acknowledgements}

We would like to thank the anonymous reviewers and our Action Editor Stefan Riezler for their comments and suggestions on this work.  We would also like to thank Daniel Beck,  Linfeng Song, Joost Bastings, Zuozhu Liu  and Yiluan Guo for their helpful suggestions. This work is supported by Singapore Ministry of Education Academic Research Fund (AcRF) Tier 2 Project MOE2017-T2-1-156. This work is also partially supported by
SUTD project PIE-SGP-AI-2018-01.

	\bibliography{tacl2018}
	\bibliographystyle{acl_natbib}
	
\end{document}

%% file: tables/table1stat.tex
\begin{table}[!t]
	\small
	\centering
	\setlength{\tabcolsep}{3pt}
	\begin{tabular}{cccc}
		\hline
		\textbf{Dataset} & \textbf{Train} & \textbf{Dev} & \textbf{Test}\\
		\hline
		AMR15 (LDC2015E86) &  ~~16,833 &  1,368 & 1,371\\
		AMR17 (LDC2017T10) &  ~~36,521 & 1,368 &  1,371 \\
		\hline
		English-Czech & 181,112 &2,656  & 2,999\\
		English-German & 226,822 &2,169  & 2,999\\
	\hline
	\end{tabular}
	\caption{The number of sentences in four datasets.}
	\label{tab:stat}
	\vspace{-3mm}
\end{table}

%% file: tables/table5amrmain.tex
\begin{table}[!t]
\centering
\small
\setlength{\tabcolsep}{2.6pt}
\scalebox{1}{
\begin{tabular}{lcccc} 
\hline
\textbf{Model} & \textbf{T} & \texttt{\#P} & \texttt{B} & \texttt{C} \\\hline
Seq2SeqB \citep{Beck2018GraphtoSequenceLU} & S  &  28,4M & 21.7 & 49.1 \\
GGNN2Seq \citep{Beck2018GraphtoSequenceLU} & S  &  28.3M & 23.3 & 50.4 \\
Seq2SeqB \citep{Beck2018GraphtoSequenceLU} & E  &  ~142M & 26.6 & 52.5 \\
GGNN2Seq \citep{Beck2018GraphtoSequenceLU} & E  &  ~141M & 27.5 & 53.5 \\
\hline 
\multirow{2}{*}{DCGCN (ours)} & S  & \bf 19.1M & 27.9 & 57.3 \\
 & E  &  92.5M & \bf 30.4 &\bf 59.6 \\
\hline
\end{tabular}}
\vspace{-2mm}
\caption{Main results on AMR17. GCNSEQ \citep{Damonte2019StructuralNE} achieves 24.5 BLEU points.  \texttt{\#P} shows the model size in terms of parameters; ``S'' and ``E'' denote single and ensemble models, respectively.}
\vspace{-1em}
\label{tab:mainamr}
\end{table}


%% file: tables/table6amrexternal.tex
\begin{table}[!t]
\centering
 \small
\setlength{\tabcolsep}{1pt}
\begin{tabular}{llcc} 
\hline
   \textbf{Model} & \textbf{External} & \texttt{B} \\
\hline 
Seq2SeqK \citep{Konstas2017NeuralAS}  & ~~~~- & 22.0 \\
GraphLSTM \citep{Song2018AGM} & ~~~~- & 23.3 \\
GCNSEQ \citep{Damonte2019StructuralNE} & ~~~~- & 24.4 \\
\hline
DCGCN(single) & ~~~~- & 25.9 \\
DCGCN(ensemble) & ~~~~- & \textbf{28.2} \\
\hline
TSP \citep{Song2016AMRtotextGA} & ~ALL &  22.4\\
PBMT \citep{Pourdamghani2016GeneratingEF}  & ~ALL & 26.9 \\
Tree2Str \citep{Flanigan2016GenerationFA}  & ~ALL & 23.0 \\
SNRG \citep{Song2017AMRtotextGW} &  ~ALL & 25.6 \\
\hline
Seq2SeqK \citep{Konstas2017NeuralAS}  & 0.2M & 27.4 \\
GraphLSTM \citep{Song2018AGM} & 0.2M & 28.2 \\
\hline
DCGCN(single) & 0.1M & 29.0 \\
DCGCN(single) & 0.2M & \textbf{31.6} \\
\hline
Seq2SeqK \citep{Konstas2017NeuralAS}  & ~~~2M & 32.3 \\
GraphLSTM \citep{Song2018AGM} & ~~~2M & 33.6 \\
Seq2SeqK \citep{Konstas2017NeuralAS}  & ~20M & 33.8 \\
\hline
DCGCN(single) & 0.3M & 33.2 \\
DCGCN(ensemble) & 0.3M & \textbf{35.3} \\
\hline
\end{tabular}
\vspace{-1mm}
\caption{Main results on AMR15 with/without external Gigaword sentences as auto-parsed data are used. The number of parameters of our single model is 18.4M}
\vspace{-4mm}
\label{tab:mainamrexternal}
\end{table}

%% file: tables/table7nmtmain.tex
\begin{table*}[!th]
\small
\centering
\setlength{\tabcolsep}{2.5pt}
\begin{tabular}{lccccccc} 
\hline
    \multirow{2}{*}{\textbf{Model}} & \multirow{2}{*}{\textbf{Type}} &
    \multicolumn{3}{c}{\textbf{English-German}} & \multicolumn{3}{c}{\textbf{English-Czech}} \\
& & \texttt{\#P} & \texttt{B} & \texttt{C} & \texttt{\#P} & \texttt{B} & \texttt{C} \\\hline
BoW+GCN \citep{Bastings2017GraphCE} & Single & - & 12.2 & - & - & ~~7.5 & - \\
CNN+GCN \citep{Bastings2017GraphCE} & Single & - & 13.7 & - & - & ~~8.7 & - \\
BiRNN+GCN \citep{Bastings2017GraphCE} & Single & - & 16.1 & - & - & ~~9.6 & - \\
\hline
PB-SMT \citep{Beck2018GraphtoSequenceLU} & Single & - & 12.8 & 43.2 & - & ~~8.6 & 36.4 \\
Seq2SeqB \citep{Beck2018GraphtoSequenceLU} & Single & ~~41.4M & 15.5 & 40.8 & ~~39.1M & ~~8.9 & 33.8 \\
GGNN2Seq \citep{Beck2018GraphtoSequenceLU} & Single & ~~41.2M & 16.7 & 42.4 & ~~38.8M & ~~9.8 & 33.3 \\\hline
DCGCN (ours) & Single & \textbf{~~29.7M} & \textbf{19.0} & \textbf{44.1} & \textbf{~~28.3M} & \textbf{12.1} & \textbf{37.1} \\
\hline
Seq2SeqB \citep{Beck2018GraphtoSequenceLU} & Ensemble & ~~~207M & 19.0 & 44.1 & ~~~195M & 11.3 & 36.4 \\
GGNN2Seq \citep{Beck2018GraphtoSequenceLU} & Ensemble & ~~~206M & 19.6 & 45.1 & ~~~194M & 11.7 & 35.9 \\
\hline
DCGCN (ours) & Ensemble & \textbf{~~~149M} & \textbf{20.5} & \bf 45.8 & \textbf{~~~142M} & \textbf{13.1} & \textbf{37.8} \\
\hline
\end{tabular}
\vspace{-1mm}
\caption{Main results on English-German and English-Czech datasets.}
\vspace{-3mm}
\label{table3}
\end{table*}

%% file: tables/table2devnum.tex
\begin{table}[!t]
	\small
	\centering
\begin{tabular}{clccc}
\hline
$Block$  & $n$ & $m$ & \texttt{B}    & \texttt{C}    \\
\hline
 \multirow{9}{*}{1} & 1 & 1 & 17.6 & 48.3 \\
 & 1 & 2 & 19.2 & 50.3 \\
 & 2 & 1 & 18.4 & 49.1 \\
 & 1 & 3 & 19.6 & 49.4 \\
 & 3 & 1 & 20.0 & 50.5 \\
 & 3 & 3 & 21.4 & 51.0 \\
 & 3 & 6 & 21.8 & 51.7 \\
 & 6 & 3 & 21.7 & 51.5 \\
 & 6 & 6 & 22.0 & 52.1 \\
\hline
\multirow{3}{*}{2} & 3 & 6 & \bf 23.5 & 53.3 \\
& 6 & 3 & 23.3 & \bf 53.4 \\
& 6 & 6 & 22.0 & 52.1   \\
\hline 
\end{tabular}
\vspace{-1mm}
\caption{The effect of the number of layers inside DCGCN sub-blocks on the AMR15 development set. }
\label{tab:devnm}
\vspace{-3mm}
\end{table}

%% file: tables/table3baseline.tex
\begin{table}[!t]
\centering
\small
\setlength{\tabcolsep}{2pt}
\begin{tabular}{lcclcc} 
\hline
\textbf{GCN} &  \texttt{B} & \texttt{C} & \textbf{GCN} &  \texttt{B} & \texttt{C}\\\hline
+RC (2) & 16.8 & 48.1 & +RC+LA (2) & 18.3 & 47.9   \\
+RC (4) &18.4  &49.6 & +RC+LA (4) &18.0 & 51.1 \\
+RC (6) &19.9  &49.7 & +RC+LA (6) &21.3 &50.8   \\
+RC (9) &\bf 21.1 &50.5  & +RC+LA (9) &\bf 22.0 &52.6   \\
+RC (10) &20.7 &\bf 50.7  & +RC+LA (10) &21.2 &\bf 52.9  \\
\hline
DCGCN1 (9) & 22.9 &	53.0  & DCGCN3 (27) & 24.8	& 54.7      \\
DCGCN2 (18) & 24.2 & 54.4 &  DCGCN4 (36) & \bf \bf 25.5	 & \bf 55.4    \\
\hline
\end{tabular}
\vspace{-1mm}
\caption{Comparisons with baselines. \text{+RC} denotes GCNs with residual connections. \text{+RC+LA} refers to GCNs with both residual connections and layer aggregations. \text{DCGCN}$i$ represents our model with $i$ blocks, containing $i \times (n+m)$ layers. The number of layers for each model is shown in parenthesis. }
\vspace{-4mm}
\label{tab:baselines}
\end{table}

%% file: tables/table2dcgcnparams.tex
\begin{table}[!t]
\centering
\small
\begin{tabular}{lcccc} 
\hline
\textbf{Model} & \texttt{D} & \texttt{\#P} & \texttt{B} &  \texttt{C}\\
\hline
DCGCN(1) & 300 & \multirow{2}{*}{10.9M}	& 20.9	& 52.0  \\
DCGCN(2) & 180 & 	& \bf 22.2	& \bf 52.3 \\
\hline
DCGCN(2) & 240	& 11.3M & 22.8 & 52.8 \\
DCGCN(4) & 180 &	11.4M &	\bf 23.4 & \bf 53.4 \\
\hline 
DCGCN(1) & 420 & 12.6M	& 22.2 &	52.4 \\
DCGCN(2) & 300 & 12.5M	& 23.8	& 53.8  \\
DCGCN(3) & 240 & 12.3M	& \bf 23.9	& \bf 54.1 \\
\hline 
DCGCN(2) & 360	& \multirow{2}{*}{14.0M}	& 24.2	& \bf 54.4 \\
DCGCN(3) & 300	& & \bf 24.4	& 54.2 \\
\hline 
DCGCN(2) & 420 & \multirow{2}{*}{15.6M}	& 24.1	& 53.7 \\
DCGCN(4) & 300	& 	& \bf 24.6	& \bf 54.8 \\
\hline 
DCGCN(3) & 420 & 18.6M  &	24.5 &	54.6 \\
DCGCN(4) & 360 & 18.4M	& \bf 25.5 & \bf 55.4 \\
\hline
\end{tabular}
\vspace{-1mm}
\caption{Comparisons of different DCGCN models under almost the  same parameter budget.}
\vspace{-3mm}
\label{tab:gcnparams}
\end{table}

%% file: tables/table4density.tex
\begin{table}[!t]
	\small
	\centering
\setlength{\tabcolsep}{3pt}
\begin{tabular}{lcc}
\hline
\bf Model & \texttt{B} & \texttt{C} \\
\hline
DCGCN4 & 25.5 &	55.4 \\
-\{4\} dense block & 24.8 &	54.9 \\
-\{3, 4\} dense blocks & 23.8 & 54.1 \\
-\{2, 3, 4\} dense blocks  & 23.2 & 53.1 \\
\hline 
\end{tabular}
\vspace{-1mm}
\caption{Ablation study for density of connections on the dev set of AMR15. -\{i\} dense block denotes removing the dense connections in the $i$-th block. }
\label{tab:density}
\end{table}

%% file: tables/table4ablation.tex
\begin{table}[!t]
 	\small
	\centering
\setlength{\tabcolsep}{3pt}
\begin{tabular}{lcc}
\hline
  \bf Model  & \texttt{B} & \texttt{C} \\
\hline
DCGCN4 & 25.5 &	55.4 \\
\hline
Encoder Modules & & \\
-Linear Combination & 23.7 & 53.2 \\
-Global Node & 24.2 & 54.6 \\
-Direction Aggregation & 24.6 & 54.6 \\
-Graph Attention  & 24.9 & 54.7 \\
-Global Node\&Linear Combination & 22.9 & 52.4 \\
\hline
Decoder Modules & & \\
-Coverage Mechanism & 23.8 & 53.0 \\
\hline 
\end{tabular}
\vspace{-1mm}
\caption{Ablation study for modules used in the graph encoder and the LSTM decoder} 
\vspace{-3mm}
\label{tab:devablation}
\end{table}

%% file: tables/table8output.tex
\begin{table}[!t]
	\small
	\centering
	\scalebox{0.72}{
	\begin{tabular}{l}
		\hline
		(s / state-01 \\
		{\color{white}0}{\color{white}0}   
		:ARG0 (p / person \\ 
		{\color{white}0}{\color{white}0}{\color{white}0}{\color{white}0}   
		:ARG0-of (h / have-org-role-91 \\
		{\color{white}0}{\color{white}0}{\color{white}0}{\color{white}0}{\color{white}0}{\color{white}0}   :ARG1 (i / intelligence  \\
		{\color{white}0}{\color{white}0}{\color{white}0}{\color{white}0}{\color{white}0}{\color{white}0}{\color{white}0}{\color{white}0}
		:mod (c / country :wiki "united$\_$states" \\
	    {\color{white}0}{\color{white}0}{\color{white}0}{\color{white}0}{\color{white}0}{\color{white}0}{\color{white}0}{\color{white}0}{\color{white}0}{\color{white}0}
	    :name (n / name :op1 "u.s.")))   \\
		{\color{white}0}{\color{white}0}{\color{white}0}{\color{white}0}{\color{white}0}{\color{white}0}
		:ARG2 (o / official)))  \\
		{\color{white}0}{\color{white}0}   
		:ARG1 (c2 / continue-01  \\
		{\color{white}0}{\color{white}0}{\color{white}0}{\color{white}0} 
		:ARG0 (p2 / person  \\
		{\color{white}0}{\color{white}0}{\color{white}0}{\color{white}0}{\color{white}0}{\color{white}0}  
		:ARG0-of (h2 / have-org-role-91 \\
		{\color{white}0}{\color{white}0}{\color{white}0}{\color{white}0}{\color{white}0}{\color{white}0}{\color{white}0}{\color{white}0}
		:ARG2 (o2 / official \\
		{\color{white}0}{\color{white}0}{\color{white}0}{\color{white}0}{\color{white}0}{\color{white}0}{\color{white}0}{\color{white}0}{\color{white}0}{\color{white}0}		
		:mod (c3 / country :wiki "north$\_$korea" \\
		{\color{white}0}{\color{white}0}{\color{white}0}{\color{white}0}{\color{white}0}{\color{white}0}{\color{white}0}{\color{white}0}{\color{white}0}{\color{white}0}{\color{white}0}{\color{white}0}		
		:name (n2 / name :op1 "north" :op2 \\
		{\color{white}0}{\color{white}0}{\color{white}0}{\color{white}0}{\color{white}0}{\color{white}0}{\color{white}0}{\color{white}0}{\color{white}0}{\color{white}0}{\color{white}0}{\color{white}0}		
		"korea"))))) \\
		{\color{white}0}{\color{white}0}{\color{white}0}{\color{white}0} 
		:ARG1 (t / trade-01\\
		{\color{white}0}{\color{white}0}{\color{white}0}{\color{white}0}{\color{white}0}{\color{white}0} 
		:ARG1 (t2 / \textbf{technology} \\
		{\color{white}0}{\color{white}0}{\color{white}0}{\color{white}0}{\color{white}0}{\color{white}0}{\color{white}0}{\color{white}0} 
		:purpose (w / weapon \\
		{\color{white}0}{\color{white}0}{\color{white}0}{\color{white}0}{\color{white}0}{\color{white}0}{\color{white}0}{\color{white}0}{\color{white}0}{\color{white}0} 
		:ARG2-of  (d / destroy-01 \\ 
		{\color{white}0}{\color{white}0}{\color{white}0}{\color{white}0}{\color{white}0}{\color{white}0}{\color{white}0}{\color{white}0}{\color{white}0}{\color{white}0}{\color{white}0}{\color{white}0} 		
		:degree (m / mass)))) \\
		{\color{white}0}{\color{white}0}{\color{white}0}{\color{white}0}{\color{white}0}{\color{white}0} 	 
		:mod (g / globe))  \\
		{\color{white}0}{\color{white}0}{\color{white}0}{\color{white}0}
		:ARG2-of (i2 / include-01\\
		{\color{white}0}{\color{white}0}{\color{white}0}{\color{white}0}{\color{white}0}{\color{white}0}
		:ARG1 (i3 / \textbf{instruct-01} \\
		{\color{white}0}{\color{white}0}{\color{white}0}{\color{white}0}{\color{white}0}{\color{white}0}{\color{white}0}{\color{white}0}		
		:ARG3 (m2 / make-01 \\
		{\color{white}0}{\color{white}0}{\color{white}0}{\color{white}0}{\color{white}0}{\color{white}0}{\color{white}0}{\color{white}0}{\color{white}0}{\color{white}0}{\color{white}0}		
		:ARG1 (m3 / missile\\
		{\color{white}0}{\color{white}0}{\color{white}0}{\color{white}0}{\color{white}0}{\color{white}0}{\color{white}0}{\color{white}0}{\color{white}0}{\color{white}0}{\color{white}0}{\color{white}0}{\color{white}0}		
		:ARG1-of (a / advanced-02)))))))  \\
		\hline
		\textbf{Reference}: u.s. intelligence officials stated that north korean \\ 
		officials  are continuing global trade in \textbf{technology} for weapons  \\ 
		of mass  destruction including \textbf{instructions} for making advanced \\ 
		missiles. \\
		\hline
		\textbf{GCN+RC}: a u.s. intelligence official stated that north korea  \\
		officials  continued the global trade for weapons of mass \\
		destruction by  making advanced missiles to make advanced\\
		 missiles. \\
		\hline
		\textbf{GCN+RC+LA}: a u.s. intelligence official stated that north  \\
		korea officials  continued global trade with weapons of mass  \\
		destruction including making advanced missiles.  \\
		\hline
		\textbf{DCGCN}: a u.s. intelligence official stated that north korea \\ 
		officials continue global trade on \textbf{technology} for weapons of  \\ 
		mass destruction including \textbf{instructions} to make advanced  \\
		missiles. \\
		\hline
	\end{tabular}}
	\caption{Example outputs.}
	\label{tab:output}
    \vspace{-3mm}
\end{table}